\documentclass[conference]{IEEEtran}
\IEEEoverridecommandlockouts
\usepackage{cite}
\usepackage{amsmath,amssymb,amsfonts}
\usepackage{algorithmic}
\usepackage{graphicx}
\usepackage{textcomp}
\usepackage{xcolor}
\def\BibTeX{{\rm B\kern-.05em{\sc i\kern-.025em b}\kern-.08em
    T\kern-.1667em\lower.7ex\hbox{E}\kern-.125emX}}
\usepackage{dsfont}
\usepackage{multirow}

\newcommand{\z}{\mathbf{z}}
\newcommand{\x}{\mathbf{x}}
\newcommand{\y}{\mathbf{y}}
\newcommand{\E}{\mathbb{E}}
\newcommand{\LL}{\mathcal{L}}
\newcommand{\cR}{ \LL_f}
\newcommand{\cI}{{J}_{\omega}}
\newcommand{\cJ}{J}
\newcommand{\sfr}{r}
\newcommand{\sR}{\mathbb{R}}
\newcommand{\ind}[1]{\mathds{1}_\mathrm{#1}} 
\usepackage{balance}

\begin{document}
\bstctlcite{IEEEexample:BSTcontrol}
\title{Generative Adversarial Networks: A Likelihood Ratio Approach\\
\thanks{This work was supported by the US National Science Foundation, Grant CIF 1513373, through Rutgers University.}
}

\author{\IEEEauthorblockN{Kalliopi Basioti}
\IEEEauthorblockA{\textit{Department of Computer Science} \\
\textit{Rutgers University}\\
Piscataway, New Jersey, USA \\
kalliopi.basioti@rutgers.edu}
\and
\IEEEauthorblockN{George V. Moustakides}
\IEEEauthorblockA{\textit{Department of Electrical and Computer Engineering} \\
\textit{University of Patras}\\
Patras, Greece \\
moustaki@ece.upatras.gr}
}

\maketitle

\begin{abstract}
We are interested in the design of generative networks. The training of these mathematical structures is mostly performed with the help of adversarial (min-max) optimization problems. We propose a simple methodology for constructing such problems assuring, at the same time, consistency of the corresponding solution. We give characteristic examples developed by our method, some of which can be recognized from other applications, and some are introduced here for the first time.  We present a new metric, the likelihood ratio, that can be employed online to examine the convergence and stability during the training of different Generative Adversarial Networks (GANs). Finally, we compare various possibilities by applying them to well-known datasets using neural networks of different configurations and sizes.
\end{abstract}


\section{Introduction}\label{sec:2}
% \section{The Generative Adversarial Problem}

The problem we are interested in, can be summarized as follows: We are given two collections of training data $\{\z_j\}$ and $\{\x_i\}$. In the first set the samples follow the origin probability density $h(\z)$ and in the second the target density $f(\x)$. The target density $f(\x)$ is considered unknown while $h(\z)$ can either be known with the possibility to produce samples $\z_j$ every time it is necessary or unknown in which case we have a second fixed training set $\{\z_j\}$. Our goal is to design a deterministic transformation $G(\z)$ so that the data $\y_j$ produced by applying the transformation $\y=G(\z)$ onto $\z_j$ follow the target density $f(\y)$.

Of course one may wonder whether the proposed problem enjoys any solution, namely, whether there indeed exists a transformation $G(\z)$ capable of transforming $\z$ into $\y$ with the former following the origin density $h(\z)$ and the latter the target density $f(\y)$. The problem of transforming random vectors has been analyzed in \cite{box1964analysis} where existence is shown under general conditions. Computing, however, the actual transformation is a completely different challenge with one of the possible solutions relying on adversarial approaches applied to neural networks.

The most well known usage of this result is the possibility to generate synthetic data that follow the unknown target density $f(\x)$. In this case $h(\z)$ is selected to be simple (e.g.~i.i.d.~standard Gaussian or i.i.d.~uniform) so that generating realizations from $h(\z)$ is straightforward. As mentioned, the adversarial approach can be applied even if the origin density $h(\z)$ is unknown provided that we have a dataset $\{\z_j\}$ with data following the origin density. 

It was \cite{goodfellow2014generative} that first introduced the idea of adversarial (min-max) optimization and demonstrated that it results in the determination of the desired transformation $G( \z)$ (consistency). Alternative adversarial approaches were subsequently suggested by \cite{arjovsky2017wasserstein, binkowski2018demystifying} and also shown to deliver the correct transformation $G(\z)$. 

In the work of \cite{nowozin2016f} a class of min-max optimizations, f-GANs, was defined to design generator/discriminator pairs. Then, \cite{liu2017approximation} defined the adversarial divergences class of objective functions, which further combined f-GANs, MMD-GAN \cite{li2017mmd}, Wasserstein GANs (WGANs) \cite{arjovsky2017wasserstein}, WGAN with Gradient Penalty \cite{gulrajani2017improved}, and entropic regularized optimal transport problems. 
%Also, they investigated under what conditions the Discriminator's class has the effect of matching generalized moments. 
Next, the work of \cite{song2020bridging} connected f-GANs, and WGANs, and later \cite{birrell2020f} generalized the results by introducing the $(f,\Gamma)-$divergencies, which allowed to bridge f-divergencies and integral probability metrics. 
A drawback of the f-GANs based families is that the derivation of a loss function requires the solution of an additional optimization problem, making it challenging to discover new GANs losses. Our work will show that our methods provide an abundance of adversarial problems capable of identifying the appropriate transformation $G(\z)$ without additional optimization to find new losses.  Instead, we will provide a simple recipe as to how we can successfully construct such problems.

Arguing along the same lines of the existing min-max formulations: We would like to optimally specify a vector transformation $G(\z)$, the generator, and a scalar function $D(\x)$, the discriminator. To achieve this, for each combination $\{G(\z),D(\x)\}$ we define the cost function
\begin{equation}\label{eq:1}
    J(G,D) = \E_{\x\sim f}[\phi(D(\x))] + \E_{\z\sim h}[\psi(D(G(\z)))]
\end{equation}
where $\phi(z),\psi(z)$ are two scalar functions of the scalar $z$. The optimum combination generator/discriminator is then identified by solving the following min-max problem
\begin{equation}\label{eq:2}
    \min_{G(\z)}\max_{D(\x)} J(G,D)
\end{equation}
We must point out that our goal is not to solve \eqref{eq:2}, but rather find a class of functions $\phi(z),\psi(z)$ so that the transformation $G(\z)$ that will come out of the solution of equation 2 is such that $\y=G(\z)$ follows the target density $f(\y)$ when $\z$ follows the origin density $h(\z)$.

If $\z$ is random following $h(\z)$ then $\y=G(\z)$ is also random and we denote with $g(\y)$ its corresponding probability density. Clearly, there exists a correspondence between transformations $G(\z)$ and densities $g(\y)$ when the density $h(\z)$ of $\z$ is fixed. Since we can write $\E_{\z\sim h}[\psi(D(G(\z)))] = \E_{\y\sim g}[\psi(D(\y))]$ this allows us to argue that the min-max problem in  \eqref{eq:2} is equivalent to
\begin{equation}\label{eq:3}
    \min_{g(\y)}\max_{D(\x)} \{ \E_{\x \sim f} [\phi(D(\x))] + \E_{\y \sim g} [\psi(D(\y))] \}
\end{equation}
It is now possible to combine the two expectations by applying a change of measure and a change of variables and equivalently write \eqref{eq:3} as follows:
\begin{multline}
\min_{g(\y)}\max_{D(\x)}\left\{\E_{\x \sim f}\big[\phi\big(D(\x)\big)\big] + \int \psi\big(D(x)\big) \frac{g(x)}{f(x)}f(x)dx\right\}\\
=\min_{g(\x)}\max_{D(\x)}\big\{\E_{\x \sim f}\big[\phi\big(D(\x)\big)\big] + 
\E_{\x \sim f}\big[r(\x)\psi\big(D(\x)\big)\big]\big\}\\
=\min_{g(\x)}\max_{D(\x)}\E_{\x \sim f}\big[\phi\big(D(\x)\big)+r(\x)\psi\big(D(\x)\big)\big]
\label{eq:4}
\end{multline}
where $r(\x)=g(\x)/f(\x)$ denotes the corresponding likelihood ratio. Since $f(\x)$ is also fixed, there is again a correspondence between $r(\x)$ and $g(\x)$, hence the previous min-max problem becomes equivalent to
\begin{equation}\label{eq:5}
    \min_{r(\x)\in \LL_f}\max_{D(\x)} \E_{\x \sim f} \big[\phi\big(D(\x)\big) + r(\x) \psi\big(D(\x)\big)\big]
\end{equation}
Here $\LL_f$ denotes the class of all likelihood ratios $r(\x)$ with respect to the density $f(\x)$, namely, all the functions $r(\x)$ that satisfy
\begin{equation}\label{eq:6}
    \LL_f = \left\{r(\x): r(\x)\geq 0, \int r(\x)f(\x)d\x = 1 \right\}
\end{equation}
Using these definitions, let us define the cost
\begin{equation}\label{eq:7}
    J(r,D) = \E_{\x\sim f}\big[\phi \big( D(\x)\big) + r(\x)\psi\big(D(\x)\big)\big]
\end{equation}
and according to \eqref{eq:4}, we are interested in the following min-max problem
\begin{equation}\label{eq:8}
    \min_{r(\x)\in \LL_f}\max_{D(\x)} J(r,D)
\end{equation}
As mentioned, our actual goal is not to solve the adversarial problem. Instead, we would like to properly identify pairs of functions $\{\phi(z),\psi(z)\}$ so that \eqref{eq:8} accepts as solution the function $r(\x)=1$. Indeed, if $r(\x)=1$ is the solution of \eqref{eq:8}, this means that $g(\x)=f(\x)$ is the solution to \eqref{eq:3} and finally the optimum $G(\x)$ obtained from \eqref{eq:1} is such that $\y=G(\x)$ follows $g(\y)=f(\y)$ which, of course, is our original objective. Even though the min-max problem in \eqref{eq:1} is what we attempt to solve, it is through \eqref{eq:8}  that we understand what its solution entails. In the next section we focus on \eqref{eq:7}, \eqref{eq:8} and propose a simple design method (recipe) for the two functions $\phi(z),\psi(z)$ that assures that the solution \eqref{eq:8} is indeed $r(\x)=1$.

Before we discuss the details of our work, we would like to summarize this paper's contribution.
\begin{itemize}
    \item We design a family of GANs problems using a likelihood ratio approach. In this class, all optimization problems have the desired property that the generator output follows the target distribution of the random vector of interest, $\x$, in other words, that the likelihood ratio of the two distributions is equal to one. 
    \item We propose a straightforward \textit{recipe} to explore the GANs family. With this methodology, we were able to identify subclasses in the GANs family characterized by specific transformations of the likelihood ratio. We identified known GANs (such as vanilla \cite{goodfellow2014generative} and Wasserstein GANs) and discovered novel ones in these subclasses.
    \item We propose a new \textit{online} metric, the likelihood ratio, for evaluating the performance of GANs during training.
    \item Our experiments provide insights for the different GANs' objective functions' behavior, with some novel objective functions performing better than the already known GANs.
\end{itemize}

\section{A class of Functions $\phi(z),\psi(z)$}

Suppose that  $\omega(r)$ is a \textit{strictly increasing and (left and right) differentiable} scalar function of the nonnegative scalar $r$, i.e.~$r\in[0,\infty)$. Denote with $\cI=\omega\big([0,\infty)\big)$ the range of values of $\omega(r)$ and let $\omega^{-1}(z)$ be the inverse function of $\omega(r)$ which exists and is defined for $z\in\cI$. Let $\rho(z)>0$ be a positive scalar function also defined for $z\in\cI$ then, using $\omega(r)$ and $\rho(z)$, we propose the following pair $\phi(z),\psi(z)$
\begin{equation}
\phi'(z)=-\omega^{-1}(z)\rho(z),~~\psi'(z)=\rho(z),
\label{eq:9}
\end{equation}
where ``$\,^\prime\,$'' denotes derivative.
Since $\omega(r)$ and $\rho(z)$ are arbitrary (provided they satisfy the strict increase and positivity constraint respectively), the class of pairs defined by \eqref{eq:9} is very rich allowing for a multitude of choices. 
We show next that \textit{any} such pair $\{\phi(z),\psi(z)\}$ gives rise to a min-max problem, as in \eqref{eq:8}, that accepts $r(\x)=1$ as its unique solution. We prove this claim in two steps. The first, involves a theorem where we consider a simplified version of the min-max problem.
%\vspace{-0.1cm}
\newtheorem{theorem}{Theorem}
\begin{theorem} \label{th:1}
Let $\omega(r),\phi(z),\psi(z)$ and $\cI$ be defined as above with the additional constraint $\psi\big(\omega(1)\big)=0$. 
Fix $r\geq0$ and consider $\phi(D)+r\psi(D)$ as a function of the scalar $D$. Then, for any $D\in\cI$, we have that
\begin{equation}\label{eq:10}
\phi(D)+r\psi(D)\leq\phi\big(\omega(r)\big)+r\psi\big(\omega(r)\big),
\end{equation}
with equality if and only if $D=\omega(r)$. 

Consider next the minimization with respect to $r$ of the maximal value in \eqref{eq:10}. It is then true that
\begin{equation}
\min_{r\geq0}\,\left\{\phi\big(\omega(r)\big)+r\psi\big(\omega(r)\big)\right\}=
\phi\big(\omega(1)\big),
\label{eq:11}
\end{equation}
with equality if and only if $r=1$.
\end{theorem}
\begin{IEEEproof}
We note that the constraint $\psi\big(\omega(1)\big)=0$ does not affect the generality of our class of functions since from \eqref{eq:9} we have that $\psi(z)$, after integration, is defined up to an arbitrary additive constant. We can always select this constant so that the constraint is satisfied. We would also like to emphasize that this constraint is needed only for the proof of this theorem and it is not necessary for the corresponding min-max problem defined in \eqref{eq:8}.

For fixed $r$, to find the maximum of $\phi(D)+r\psi(D)$ we consider the derivative with respect to $D$ which, using \eqref{eq:9}, takes the form
\begin{equation}\label{eq:12}
\phi'(D)+r\psi'(D)=\big(r-\omega^{-1}(D)\big)\rho(D).
\end{equation}
The strict increase of $\omega(r)$ is inherited by its inverse function $\omega^{-1}(z)$ which, combined with the positivity of $\rho(z)$, implies that the previous expression has the same sign as $r-\omega^{-1}(D)$ or $\omega(r)-D$. Consequently $D=\omega(r)$ is the \textit{only critical point} of $\phi(D)+r\psi(D)$ which  is a global maximum. Of course there are possibilities for extrema at the two end points of $\cI$ but they can only be (local) minima.

Let us now focus on the resulting function $\phi\big(\omega(r)\big)+r\psi\big(\omega(r)\big)$. Taking its derivative with respect to $r$ yields
\begin{multline}
\!\!\!\left\{\phi\big(\omega(r)\big)+r\psi\big(\omega(r)\big)\right\}'=\left\{\phi'\big(\omega(r)\big)+r\psi'\big(\omega(r)\big)\right\}\omega'(r)\\
+\psi\big(\omega(r)\big)=
\psi\big(\omega(r)\big),
\label{eq:13}
\end{multline}
where the last equality is due to the specific definition of the two functions $\phi(z),\psi(z)$ in \eqref{eq:9}. Since $\psi'(z)=\rho(z)>0$, this implies that $\psi(z)$ is strictly increasing, being also the integral of $\rho(z)$ it is continuous in $z$. If we combine this property with the strict increase and continuity (as a result of left and right differentiability) of $\omega(r)$ we conclude that $\psi\big(\omega(r)\big)$ is also strictly increasing and continuous in $r$. We recall that $\psi(z)$ is selected to satisfy $\psi\big(\omega(1)\big)=0$, consequently for $r=1$ the function $\phi\big(\omega(r)\big)+r\psi\big(\omega(r)\big)$ has a unique minimum which is global and no other critical points. Of course it can still exhibit extrema at $r=0$ and/or $r\to\infty$ but they can only be (local) maxima.
\end{IEEEproof}

A consequence of Theorem\,\ref{th:1} is the next corollary, which constitutes the second and final step in proving that the adversarial problem defined in \eqref{eq:8} has as unique solution the function $r(\x)=1$.

\newtheorem{corollary}{Corollary}
\begin{corollary}  \label{cor:1}
If the functions $\phi(z),\psi(z)$ satisfy \eqref{eq:9} and $\omega(r)$ is strictly increasing and left and right differentiable, then in the adversarial problem defined in \eqref{eq:8} the maximizer is $D(\x)=\omega\big(r(\x)\big)$ and the minimizer is $r(\x)=1$, while the resulting min-max value is equal to
\begin{multline}
\min_{r(\x)\in\cR}\max_{D(\x)}\,\E_{\x\sim f}\left[\phi\big(D(\x)\big)+r(\x)\psi\big(D(\x)\big)\right]\\
=\phi\big(\omega(1)\big)+\psi\big(\omega(1)\big).
\label{eq:14}
\end{multline}
\end{corollary}

\begin{IEEEproof}
First, we observe that
\begin{multline}
\E_{\x \sim f}\left[\phi\big(D(\x)\big)+r(\x)\psi\big(D(\x)\big)\right]\\
=\E_{\x \sim f}\left[\phi\big(D(\x)\big)+r(\x)\tilde{\psi}\big(D(\x)\big)\right]+\psi\big(\omega(1)\big)
\label{eq:15}
\end{multline}
with the last equality being true since $\E_{\x \sim f}[r(\x)]=1$ and where $\tilde{\psi}(z)=\psi(z)-\psi\big(\omega(1)\big)$. We start with the maximization problem. Since $D(\x)$ is a function of $\x$ we have
\begin{multline}
\max_{D(\x)}\,\E_{\x\sim f}\left[\phi\big(D(\x)\big)+r(\x)\tilde{\psi}\big(D(\x)\big)\right]\\
=\E_{\x\sim f}\left[\max_{D(\x)}\left\{\phi\big(D(\x)\big)+r(\x)\tilde{\psi}\big(D(\x)\big)\right\}\right].
\label{eq:16}
\end{multline}
The maximization under the expectation can be performed for each fixed $\x$. However, when we fix $\x$ then $r(\x)$ becomes a constant and the result of the maximization depends only on the actual value of $r(\x)$. This suggests that we can limit ourselves to functions of the form $D(\x)=D\big(r(\x)\big)$. After this observation we can drop the dependence on $\x$ and perform, equivalently, the maximization
$
\max_{D}\big\{\phi\big(D(r)\big)+r\tilde{\psi}\big(D(r)\big)\big\}
$ 
for each fixed $r$. The pair $\{\phi(z),\tilde{\psi}(z)\}$ satisfies the assumptions of Theorem\,\ref{th:1}, therefore maximization is achieved for $D(r)=\omega(r)$. This implies that
\begin{multline}\label{eq:17}
\!\!\max_{D(\x)}\,\E_{\x \sim f}\left[\phi\big(D(\x)\big)+r(\x)\psi\big(D(\x)\big)\right]=\\
\E_{\x \sim f}\left[\phi\big(\omega\big(r(\x)\big)\big)+r(\x)\tilde{\psi}\big(\omega\big(r(\x)\big)\big)\right]+\psi\big(\omega(1)\big).\!\!
\end{multline}
We can now continue in a similar way for the minimization problem. Specifically
\begin{multline}
\min_{r(\x)\in\cR}\max_{D(\x)}\,\E_{\x \sim f}\left[\phi\big(D(\x)\big)+r(\x)\tilde{\psi}\big(D(\x)\big)\right]\\
=\min_{r(\x)\in\cR}\E_{\x\sim f}\left[\phi\big(\omega\big(r(\x)\big)\big)+r(\x)\tilde{\psi}\big(\omega\big(r(\x)\big)\big)\right]\\
\geq\E_{\x \sim f}\left[\min_{r(\x)\in\cR}\left\{\phi\big(\omega\big(r(\x)\big)\big)+r(\x)\tilde{\psi}\big(\omega\big(r(\x)\big)\big)\right\}\right]\\
\geq\E_{\x \sim f}\left[\min_{r}\left\{\phi\big(\omega(r)\big)+r\tilde{\psi}\big(\omega(r)\big)\right\}\right]=
\phi\big(\omega(1)\big)
\label{eq:18}
\end{multline}
with the last inequality being true since the minimization that follows is unconstrained and the last equality being a consequence of Theorem\,\ref{th:1}. The final lower bound is clearly attained by $\sfr(\x)=1$, which is also a legitimate solution of the constrained minimization, since $\sfr(\x)=1$ belongs to the class $\cR$ of likelihood ratios. Consequently $\sfr(\x)=1$ is the solution to the min-max problem. Returning to the original min-max setup with $\psi(z)$ replacing $\tilde{\psi}(z)$, we can clearly see that it satisfies \eqref{eq:14}. This completes the proof.
\end{IEEEproof}

\newtheorem{remark}{Remark}
\begin{remark}
The  adversarial  problem  is  defined  with  the help  of  the  two  functions $\phi(z), \psi(z)$ which,  according to \eqref{eq:9},  can  be  obtained  by  integrating  the  corresponding derivatives.   However,  this  integration  might  not  always be  possible,  analytically.   As  we  will  have  the  chance  to confirm in Section \ref{sec:4}, in an actual optimization algorithm (e.g.  of  gradient  type)  that  solves \eqref{eq:2},  the  exact  form  of $\phi(z), \psi(z)$ is not necessary.   Instead,  what is required is their derivatives which are analytically available from \eqref{eq:9}.
\end{remark}

We must emphasize that there already exists the significant work by \cite{nowozin2016f} that addresses a similar problem as our current work, namely the definition of a class of min-max optimizations that can be used to design the generator/discriminator pair. The class in \cite{nowozin2016f} is defined in terms of a convex function $f(\sfr)$ which can be shown to correspond to the outcome of our maximization, namely the function $\phi\big(\omega(\sfr)\big)+\sfr\psi\big(\omega(\sfr)\big)$. This establishes a one-to-one correspondence between the two methods under the ideal (non data-driven) setup. However, we believe that, our approach enjoys certain significant advantages: 

First, the definition of the two functions $\phi(z),\psi(z)$ in \eqref{eq:3} is straightforward while in \cite{nowozin2016f} requires the solution of an optimization problem. 

Second, in our case we have complete control over the result of the maximization problem that defines the discriminator. In other words we can decide what transformation $\omega(\sfr)$ of the likelihood ratio $\sfr$, the discriminator must estimate. In \cite{nowozin2016f} such flexibility does not exist. 

Controlling the function we estimate with the discriminator plays a significant role in the implementation of our method. Indeed when we use a neural network to approximate the optimum discriminator, this affects the overall quality of the resulting generator/discriminator pair. We should also note that there are important applications in Statistics where one is interested in estimating only the transformation of the likelihood ratio, with the most common cases being the likelihood ratio itself, its logarithm (log-likelihood ratio), or the ratio $\frac{\sfr}{1+\sfr}$ which plays the role of the posterior probability between two densities. In other words, there are applications where one is interested only in the ``max'' part of the min-max problem. In fact, in the next section we give examples of various choices of $\omega(\sfr)$ and mention problems where the discriminator function becomes the actual target and not the generator.

\subsection{Subclasses of the GAN Family}\label{sec:subclasses}
\centerline{\underline{\textbf{Subclass A}: $\omega(r)=r^\alpha$}} 

The first subclass we examine is the simplest one, consisting of just powers of the likelihood ratio. We should mention that this is the first work proposing objective functions from this class. To find the pairs $\{\phi(z),\psi(z)\}$ we proceed as follows.

We have that $\omega^{-1}(z)=z^{\frac{1}{\alpha}}$ and $\cI=[0,\infty)$. According to \eqref{eq:9}, for $z\in[0,\infty)$ we must define $\phi'(z)=-z^{\frac{1}{\alpha}}\rho(z),~~\psi'(z)=\rho(z).$  The likelihood ratio with respect to the Discriminator function and the parameter $a$ is $r = D^{-a}.$
The following examples can be shown to satisfy these equations.

A1)~If we select $\rho(z)=z^{\beta}$, with $\beta\neq-1,-1-\frac{1}{\alpha}$, this yields $\phi(z)=-\frac{z^{1+\frac{1}{\alpha}+\beta}}{1+\frac{1}{\alpha}+\beta}$ and $\psi(z)=\frac{z^{1+\beta}}{1+\beta}$. For $\beta=-1$, $\rho(z)=z^{-1}$, $\phi(z)=-\alpha z^{\frac{1}{\alpha}}$, $\psi(z)=\log z$. For $\beta=-1-\frac{1}{\alpha}$, $\rho(z)=z^{-1-\frac{1}{\alpha}}$, $\phi(z)=-\log z$, $\psi(z)=-\alpha z^{-\frac{1}{\alpha}}$.

A2)~If we select $\alpha=1$, $\rho(z)=\frac{1}{(1+z)}$ then, $\phi(z)=-(1+z)$ and $\psi(z)=-(1+z^{-1})$.

A3)~If we select $\alpha=1$, $\rho(z)=\frac{1}{(1+z)z}$ then, $\phi(z)=-\log(1+z)$ and $\psi(z)=-\log(1+z^{-1})$.

For the particular selection $\omega(r)=r$ (corresponding to $\alpha=1$) we can show that the resulting cost is equivalent to the Bregman cost \cite{bregman1967relaxation}. In fact there is a one-to-one correspondence between our $\rho(z)$ function and the function that defines the Bregman cost. This correspondence however is lost once we switch to a different $\alpha$ or a different $\omega(r)$ function, suggesting that the proposed class of pairs $\{\phi(z),\psi(z)\}$, is far richer than the class induced by the Bregman cost.

We should mention that in A1) the selection $\alpha=1, \beta = 0$ is known as the mean square error criterion and if we apply
only the maximization problem then this corresponds to a
likelihood ratio estimation technique proposed in the literature by \cite{sugiyama2012density}. We will refer to this case as the \textit{MSE} GAN.
\vskip0.2cm
\centerline{\underline{\textbf{Subclass B}: $\omega(r)=\alpha^{-1}\log r$}}

This subclass considers one of the most popular transformations of the likelihood ratio, the log-likelihood ratio. As in the first subclass,  for the first time, the next examples are presented. They can be used either under a min-max setting, for the determination of the generator/discriminator pair, or under a pure maximization setting for the direct estimation of the log-likelihood ratio function $\log r(\x)$.

We have $\omega^{-1}(z)=e^{\alpha z}$ and $\cI=\sR$. As before $\rho(z)$ must be strictly positive and, according to \eqref{eq:9}, for all real $z$ we must define $\phi'(z)=-e^{\alpha z}\rho(z),~\psi'(z)=\rho(z)$. Then the likelihood ratio is given by $r=e^{aD}$.
The following examples satisfy these equations.

B1)~If $\rho(z)=e^{-\beta z}$ with $\beta\neq0,\alpha$, this produces
$
\phi(z)=-\frac{e^{(\alpha-\beta)z}}{\alpha-\beta},~~\psi(z)=-\frac{e^{-\beta z}}{\beta}.
$
If $\beta=0$ then
$\rho(z)=1$, $\phi(z)=-\frac{e^{\alpha z}}{\alpha}$, $\psi(z)=z$. If $\beta=\alpha$ then $\rho(z)=e^{-\alpha z}$, $\phi(z)=-z$ and $\psi(z)=-\frac{e^{-\alpha z}}{\alpha}$. 
We call the $\alpha = 1,\beta=0.5$ case the \textit{Exponential} GAN.

B2)~If $\alpha=1$, $\rho(z)=\frac{1}{1+e^z}$ then, $\phi(z)=-\log(1+e^z)$ and $\psi(z)=-\log(1+e^{-z})$.

\vskip0.2cm
\centerline{\underline{\textbf{Subclass C}: $\omega(r)=\frac{r}{r+1}$}}

As we already mentioned, this is another important transform of the likelihood ratio. Interestingly, in this subclass belongs the first introduced GAN \cite{goodfellow2014generative} the \textit{Cross Entropy} GAN.

When $\omega(r)=\frac{r}{r+1}$ we have $\omega^{-1}(z)=\frac{z}{1-z}$ and $\cI=[0,1]$.
For $\rho(z)>0,z\in[0,1]$ we must define the functions $\phi(z),\psi(z)$ according to \eqref{eq:9}
$\phi'(z)=-\frac{z}{1-z}\rho(z),~~\psi'(z)=\rho(z)$. In this case the likelihood ratio is $r = \frac{D}{1-D}$.  The next set of examples can be seen to satisfy these equations.

C1)~If we select $\rho(z)=\frac{1}{z}$, this yields $\phi(z)=\log(1-z)$ and $\psi(z)=\log z$. 

C2)~Selecting $\rho(z)=(1-z)^\alpha$, with $\alpha\neq0,-1$, yields $\phi(z)=-\frac{1}{1+\alpha}(1-z)^{\alpha+1}+\frac{1}{\alpha}(1-z)^\alpha$ and $\psi(z)=-\frac{1}{1+\alpha}(1-z)^{1+\alpha}$. For $\alpha=0$, we have $\rho(z)=1$ and $\phi(z)=z+\log(1-z)$, $\psi(z)=z$, while for $\alpha=-1$ we have $\rho(z)=\frac{1}{1-z}$ and $\phi(z)=-\log(1-z)-\frac{1}{1-z}$, $\psi(z)=-\log(1-z)$.

In C1) we recognize the functions used in the original article by \cite{goodfellow2014generative}. C2) appears for the first time. 

\vskip0.2cm
\centerline{\underline{\textbf{Subclass D}:  $\omega(r)=\mathrm{sign}(\log r)$}}

This is a special case of $\omega(r)$ with the corresponding function not being strictly increasing. It turns out that we can still come up with optimization problems, two of which are known and used in practice, by considering $\omega(r)$ as a \textit{limit} of a sequence of strictly increasing functions.

\textit{Monotone Loss:} As a first approximation we propose $\mathrm{sign}(z)\approx\tanh(\frac{c}{2}z)$ where $c>0$ a parameter. We note that $\lim_{c\to\infty}\tanh(\frac{c}{2}z)=\mathrm{sign}(z)$. Using this approximation we can write
\begin{equation}
\mathrm{sign}(\log r)\approx\tanh\left(\frac{c}{2}\log r\right)=\frac{r^c-1}{r^{c}+1}=\omega(r).
\label{eq:19}
\end{equation}
As we mentioned, we have exact equality for $c\to\infty$. Let us perform our analysis by assuming that $c$ is finite. We note that $\omega^{-1}(z)=(\frac{1+z}{1-z})^{\frac{1}{c}}$ and $\cI=[-1,1]$. Consequently, if $\rho(z)>0$ for $z\in[-1,1]$, we must define $\textstyle\phi'(z)=-\left(\frac{1+z}{1-z}\right)^{\frac{1}{c}}\rho(z),~\psi'(z)=\rho(z)$.
\textit{We notice that since $c\to\infty$ we cannot find the likelihood ratio in terms of the Discriminator function.}

D1)~If we let $c\to\infty$ in order to converge to the desired sign function, this yields $\phi'(z)=-\rho(z)$ and $\psi'(z)=\rho(z)$. This suggests that $\phi(z)=-\int^z\rho(x)dx$ is decreasing and $\psi(z)=\int^z\rho(x)dx=-\phi(z)$ is increasing. In fact any strictly increasing function $\psi(z)$ can be adopted provided we select $\phi(z)=-\psi(z)$.

There is a popular combination that falls under Case D1). In particular, the selection $\psi(z)=z=-\phi(z)$ reminds us of Wasserstein GAN \cite{arjovsky2017wasserstein}, with two differences, in our case $z$ should lie in $[-1,1]$ and the discriminator is not constrained to be a Lipschitz function.

\textit{Hinge Loss:} 
As a second approximation we use the expression $\mathrm{sign}(z)\approx \mathrm{sign}(z)|z|^{\frac{1}{c}},~~c>0$, which is strictly increasing, continuous and converges to $\mathrm{sgn}(z)$ as $c\to\infty$. This suggests that
\begin{equation}
\mathrm{sign}(\log r)\approx \mathrm{sign}(\log r)|\log r |^{\frac{1}{c}}=\omega(r),
\label{eq:20}
\end{equation}
and $\omega^{-1}(z)=e^{z^{c}}$. Since $\omega(r)$ can assume any real value we conclude that $\cI=\sR$ which, clearly, differs from the previous approximation where we had $\cI=[-1,1]$. If $\rho(z)>0,z\in\sR$ then, according to \eqref{eq:9} we must define $\phi'(z)=-e^{z^{c}}\rho(z),~\psi'(z)=\rho(z)$. We present the following case that leads to a very well known pair from a completely different application.

D2)~If we select $\psi'(z)=\rho(z)=\{e^{-|z|^{\frac{1}{c}}}+ \ind{z<-1}\}>0$ then $\phi'(z)=-e^{z^{\frac{1}{c}}}\{e^{-|z| ^{\frac{1}{c}}}+\ind{z<-1}\}$. If we now let $c\to\infty$, we obtain the limiting form for the derivatives which become $\psi'(z)=-\ind{z<1}$ and $\phi'(z)=\ind{z>-1}$. By integrating we arrive at $\phi(z)=-\max\{1+z,0\}$ and $\psi(z)=-\max\{1-z,0\}$.
The cost based on this particular pair is called the \textit{hinge loss} \cite{tang2013deep} and it is very popular in binary classification where one is interested only in the maximization problem. The corresponding method is known to exhibit an overall performance which in practice is considered among the best \cite{rosasco2004loss,janocha2017loss}. Here, as in \cite{zhao2016energy}, we propose the hinge loss as a means to perform adversarial optimization for the design of the generator $G(\x)$.

This completes our presentation of examples. However, we must emphasize, that these are only a few illustrations of possible pairs $\{\phi(z),\psi(z)\}$ one can construct. Indeed combining, as dictated by \eqref{eq:9}, any strictly increasing function $\omega(r)$ with any positive function $\rho(z)$ generates a legitimate pair $\{\phi(z),\psi(z)\}$ and a corresponding min-max problem \eqref{eq:8} that enjoys the desired solution $r(\x)=1$. Finally, in Table\,\ref{table:1} we summarize some of the GANs presented above which we will use later in our experiments.

\begin{table}[t]
\caption{Optimization problems for GANs}
% \vskip-0.2cm
\label{gansLoss}
\begin{center}
\begin{tabular}{|l|l|l|l|l|}
\hline
\multicolumn{1}{|c|}{\bf GAN}&\multicolumn{1}{|c|}{$\phi(z)$} &\multicolumn{1}{|c|}{$\psi(z)$}&\multicolumn{1}{|c|}{$\cI$}
\\\hline
A1a & $\displaystyle -z$  & $\displaystyle \log(z)$   &  $[0,\infty)$\\\hline
A1b &$\displaystyle -\log(z)$  & $\displaystyle -z^{-1}$   &  $[0,\infty)$\\\hline
A2 &$\displaystyle -(1+z)$   &$\displaystyle -(1+z^{-1})$   &  $[0,\infty)$\\\hline
A3 &$\displaystyle -\log(1+z)$   & $\displaystyle -\log(1+z^{-1})$   &  $[0,\infty)$\\\hline
MSE &$\displaystyle -0.5z^2$ & $\displaystyle z$ &  $[0,\infty)$\\\hline
B1a &$\displaystyle -e^z$&$\displaystyle e^{z}$  & $\sR$\\\hline
B1b &$\displaystyle -z$ & $\displaystyle-e^{-z}$  & $\sR$\\\hline
Exponential &$\displaystyle -e^{0.5z}$ & $\displaystyle-e^{-0.5z}$  & $\sR$ \\\hline
B2 &$\displaystyle -\log(1+e^{z})$ & $\displaystyle -\log(1+e^{-z})$ & $\sR$
\\\hline
Cross Entropy &$\displaystyle \log(1-z)$ & $\displaystyle \log(z)$\  & $[0,1]$\\\hline
C2 &$\displaystyle z + \log(1-z)$ & $\displaystyle z$\  & $[0,1]$
\\\hline
Hinge &$\displaystyle -(1+z)_+$ & $\displaystyle -(1-z)_+$ & $\sR$\\\hline
Wasserstein &$\displaystyle z$ & $\displaystyle -z$ &$\sR$\\\hline
\end{tabular}
\end{center}
\label{table:1}
\end{table}

\begin{table*}[h!]
\tabcolsep=4pt
\caption{KID and FID Scores}
% \vskip-0.2cm
\label{KID}
\begin{center}
\begin{tabular}{|l|c|c|c|c|c|c|c|c|}
% \hline
\hline
\multirow{3}{*}{\textbf{GANs}} & \multicolumn{2}{c|}{\textbf{CARS}} & \multicolumn{2}{c|}{\textbf{CELEBA}} & \multicolumn{2}{c|}{\textbf{CIFAR10}} & \multicolumn{2}{c|}{\textbf{MNIST}} \\ 
\cline{2-9} 
& \textbf{KID} & \textbf{FID} & \textbf{KID} & \textbf{FID} & \textbf{KID} & \textbf{FID} & \textbf{KID} & \textbf{FID} \\  
\cline{2-9} 
\multicolumn{1}{|c|}{} & \multicolumn{1}{c|}{$\!\times\!10^{-\!6}\!\!\pm\!\!\times\!10^{-\!12}$} & \multicolumn{1}{c|}{} & \multicolumn{1}{l|}{$\!\times\!10^{-\!6}\!\!\pm\!\!\times\!10^{-\!12}$} & \multicolumn{1}{c|}{} & \multicolumn{1}{l|}{$\!\times\!10^{-\!6}\!\!\pm\!\times\!10^{-\!12}$} & \multicolumn{1}{c|}{} & \multicolumn{1}{c|}{$\!\times\!10^{-\!4}\!\!\pm\!\!\times\!10^{-\!8}$} & \multicolumn{1}{c|}{} \\ \hline
A1a & $\ \ 4.03\pm 2.45$ & $23.30 \pm 0.10$ & $\ \ 2.57 \pm\ \  5.05$ & $\ \ 7.53 \pm 0.02$ & $2.75 \pm\ \  8.47$ & $9.67 \pm 0.01$ & $7.80\pm 5.86$ & $2.18 \pm0.01$ \\ \hline
A1b & $\ \ 3.33\pm 1.89$ & $24.22 \pm 0.22$ & $\ \ 2.67 \pm\ \  8.35$ & $\ \ 7.68 \pm 0.04$ & $2.36 \pm\ \  5.86$ & $9.69 \pm 0.04$ & $6.30 \pm 4.32$ & $2.13 \pm 0.01$ \\ \hline
A2 & $\ \ 4.36\pm 3.18$ & $24.40 \pm 0.04$ & $\ \ 3.38 \pm\ \  9.68$ & $\ \ 7.62 \pm 0.04$ & $3.09 \pm\ \  8.35$ & $9.53 \pm 0.04$ & $8.79 \pm 4.65$ & $2.17 \pm 0.01$ \\ \hline
A3 & $\ \ 4.42\pm 2.49$ & $23.64 \pm 0.23$ & $\ \ 7.29 \pm\ \  5.48$ & $\ \ 8.50 \pm 0.06$ & $2.40 \pm\ \  4.95$ & $9.72 \pm 0.04$ & $8.53 \pm 7.35$ & $2.15 \pm 0.01$ \\ \hline
MSE & $\ \ 3.79\pm 1.71$ & $23.99 \pm 0.04$ & $\ \ 2.55 \pm\ \  9.34$ & $\ \ 7.66 \pm 0.07$ & $2.09 \pm\ \  5.63$ & $9.61 \pm 0.03$ & $8.64\pm 7.51$ & $2.15 \pm 0.01$ \\ \hline
B1a & $\ \ 6.17 \pm 3.62$ & $23.60 \pm 0.12$ & $\ \ 5.56 \pm\ \  9.78$ & $\ \ 8.06 \pm 0.02$ & $2.35 \pm\ \  5.52$ & $9.67 \pm 0.05$ & $8.05 \pm 3.30$ & $2.13 \pm 0.01$ \\ \hline
B1b & $\ \ 5.24 \pm 2.27$ & $23.91 \pm 0.07$ & $\ \ 7.32 \pm\ \  1.18$ & $\ \ 8.07 \pm 0.06$ & $2.63 \pm\ \  8.42$ & $9.67 \pm 0.04$ & $6.85 \pm 3.71$ & $2.18 \pm 0.01$ \\ \hline
Exponential & $\ \ 5.49 \pm 2.86$ & $23.52 \pm 0.06$ & $10.05 \pm\ \  6.10$ & $\ \ 9.15 \pm 0.03$ & $2.81 \pm\ \  8.34$ & $9.79 \pm 0.03$ & $6.88\pm 3.41$ & $2.13 \pm 0.01$ \\ \hline
B2 & $\ \ 7.06 \pm 3.40$ & $23.46 \pm 0.12$ & $12.48 \pm\ \  7.38$ & $\ \ 9.39 \pm 0.06$ & $2.47 \pm\ \  5.42$ & $9.79 \pm 0.07$ & $8.82\pm 3.69$ & $2.09 \pm 0.01$ \\ \hline
Cross-Entropy & $13.09\pm 8.95$ & $25.39 \pm 0.16$ & $\ 3.53 \pm 12.65$ & $\ \ 7.53 + 0.03$ & $2.53 \pm\ \  5.58$ & $9.72 \pm 0.09$ & $6.16 \pm 5.37$ & $2.08 \pm 0.01$ \\ \hline
C2 & $\ \ 8.45\pm 2.92$ & $24.36 \pm 0.12$ & $\ 3.11 \pm\ \ 9.72$ & $\ \ 7.47 \pm 0.01$ & $4.17 \pm 11.03$ & $9.75 \pm 0.03$ & $7.72 \pm  3.36$ & $2.10 \pm 0.01$ \\ \hline
Hinge & $\ \ 4.97\pm 3.50$ & $22.88 \pm 0.14$ & $26.65 \pm 16.81$ & $10.91 \pm 0.05$ & $3.97 \pm\ \  9.42$ & $9.99 \pm 0.03$ & $8.33 \pm 4.16$ & $2.16 \pm 0.01$ \\ \hline
Wasserstein & $\ \ 4.92\pm 2.70$ & $23.99 \pm 0.04$ & $20.59 \pm 28.38$ & $10.99 \pm 0.04$ & $1.74 \pm\ \  5.22$ & $9.74 \pm 0.02$ & $7.33 \pm 3.33$ & $2.16 \pm 0.01$ \\ \hline
\end{tabular}
\end{center}
\label{table:2}
\end{table*}
\section{Data-Driven Setup and Neural Networks}\label{sec:4}
%\vskip-0.2cm
Let us now consider the data-driven version of the problem. As mentioned, the target density $f(\x)$ is unknown. Instead we are given a collection of realizations $\{\x_i\}$ that follow $f(\x)$ and a second collection $\{\z_j\}$ that follows the origin density $h(\z)$. These data constitute our training set. Regarding the second set $\{\z_j\}$ it can either become available ``on the fly'' when $h(\z)$ is known by generating realizations every time they are needed, or it can be considered fixed from the start exactly as $\{\x_i\}$, if $h(\z)$ is also unknown.
As we pointed out in Section\,\ref{sec:2}, we are interested in designing a generator $G(\z)$ so that when we apply it onto the data $\z_j$, that is, $\y_j=G(\z_j)$ the resulting $\y_j$ will follow a density that matches the target density $f(\x)$.
Since we are now considering the data-driven version of the problem, we are going to limit $G(\z),D(\x)$ to be the outputs of corresponding neural networks. Therefore the generator is replaced by $G(\z,\theta)$ while the discriminator by $D(\x,\vartheta)$ where $\theta,\vartheta$ summarize the parameters of the two neural networks.

Once we have selected our favorite $\omega(r)$ and $\rho(z)$ functions we can compute from \eqref{eq:9} the functions $\phi(z),\psi(z)$ that enter into the min-max problem defined in \eqref{eq:2}. This problem, after limiting the generator and discriminator to neural networks, can be rewritten as follows
\begin{multline}
\!\!\!\!\min_{\theta}\max_{\vartheta}\cJ(\theta,\vartheta)=
\min_{\theta}\max_{\vartheta}\big\{\E_{\x \sim f}\big[\phi\big(D(\x,\vartheta)\big)\big]\\
+\E_{\z \sim h}\big[\psi\big(D(G(\z,\theta),\vartheta)\big)\big]\big\}.\!\!
\label{eq:3001}
\end{multline}
If $\theta_o,\vartheta_o$ are the corresponding optimum parameter values, and the structure of the two networks is sufficiently rich, we expect that $G(\z,\theta_o),D(\x,\vartheta_o)$ will approximate the optimum functions $\displaystyle D(\x),G(\z)$ of the ideal problem in \eqref{eq:2} respectively. In particular for $\theta_o$, the generator $G(\z,\theta_o)$, whenever applied onto any $\z_j$ that follows $h(\z)$, it will result in a $\y_j=G(\z_j,\theta_o)$ that follows a density which is expected to be close to the target density $f(\y)$.
\begin{remark}
 When replacing $D(\x),G(\z)$ with neural networks we must take special care of the corresponding outputs. Basically, we must guarantee that they are of the correct form.
 This is particularly important in the case of the scalar output $D(\x,\vartheta)$ of the discriminator. We recall that the optimum discriminator is $D(\x)=\omega\big(r(\x)\big)$. This implies that we need to assure that $D(\x,\vartheta)$ takes values in $\cI$ (the range of $\omega(r)$). Consequently, we must apply the proper nonlinearity in the output of the discriminator that will guarantee this fact.
\end{remark}
\section{Experiments}\label{sec:5}

In this section, we want to examine the performance of the GANs objectives presented in Table\,\ref{gansLoss} for different datasets.
For that reason we tested their performance on four different datasets, namely MNIST \cite{lecun1998gradient}, CelebA \cite{liu2015deep}, CIFAR-10 datasets \cite{krizhevsky2009learning}, and Stanford Cars \cite{krause20133d}. 
We recall that GANs are notorious for their nonrobust behavior \cite{bengio2012practical, creswell2018generative, mescheder2017numerics}. For stabilizing the training process, we used the maximum gradient-penalty methodology, which was generalized to a class of Lipschitz GANs in \cite{zhou2019lipschitz}. 

For the generator, we used a four-layer neural network where the first layer is linear and the remaining deconvolutional; with ReLU activation functions between the layers except the final layer where we used the hyperbolic tangent function since the output is an image with pixel values in the range $[-1,1]$. The image range was changed from $[0,1]$ to $[-1,1]$ because the hyperbolic tangent speeds the convergence of GANs compared to sigmoid (used for the $[0,1]$ range).  The generator input is a standard i.i.d.~normal vector with dimension 64 for MNIST and 128 for CelebA, CIFAR-10, and Stanford Cars. 

For the discriminator, we used a four-layer neural network with three convolutional layers followed by a linear layer. We applied Leaky ReLUs between the layers except for the final layer, where we adopted proper functions based on the range $\cI$. For the training of the two neural networks we applied the Adam algorithm \cite{kingma2014adam} with $\beta_1 = 0.5$, $\beta_2 = 0.9$, learning rate $10^{-4}$ and batch size 50 for MNIST and 64 for CelebA and CIFAR-10.

In Table\,\ref{table:2} we present the final attained Frechet Inception Distances (FID) \cite{heusel2017gans} and Kernel Inception Distances (KID) \cite{binkowski2018demystifying} scores after training for $2\times 10^5$ Generator iterations. We make this distinction since the Discriminator performs five critic iterations for each Generator iteration. Our results indicate that some losses have inferior performance compared with the others.  The losses that seem to attain poor FID/KID scores are the B2, Cross-Entropy, and C2 GANs in Stanford Cars and Exponential, B2, Hinge, and Wasserstein GANs in CelebA dataset. On the other hand, the A1a, A1b, A2, and Mean Square have excellent performance concerning all examined datasets. Furthermore, we notice that for simpler datasets (MNIST), the various GANs losses perform similarly. Still, when the dataset complexity increases (CelebA, CIFAR-10 Stanford Cars) and a small number of examples are available (only $\sim8000$ samples available in the Stanford Cars dataset), visible performance variations emerge for different GANs.
\begin{figure}[b]
\centering\includegraphics[width=\hsize]{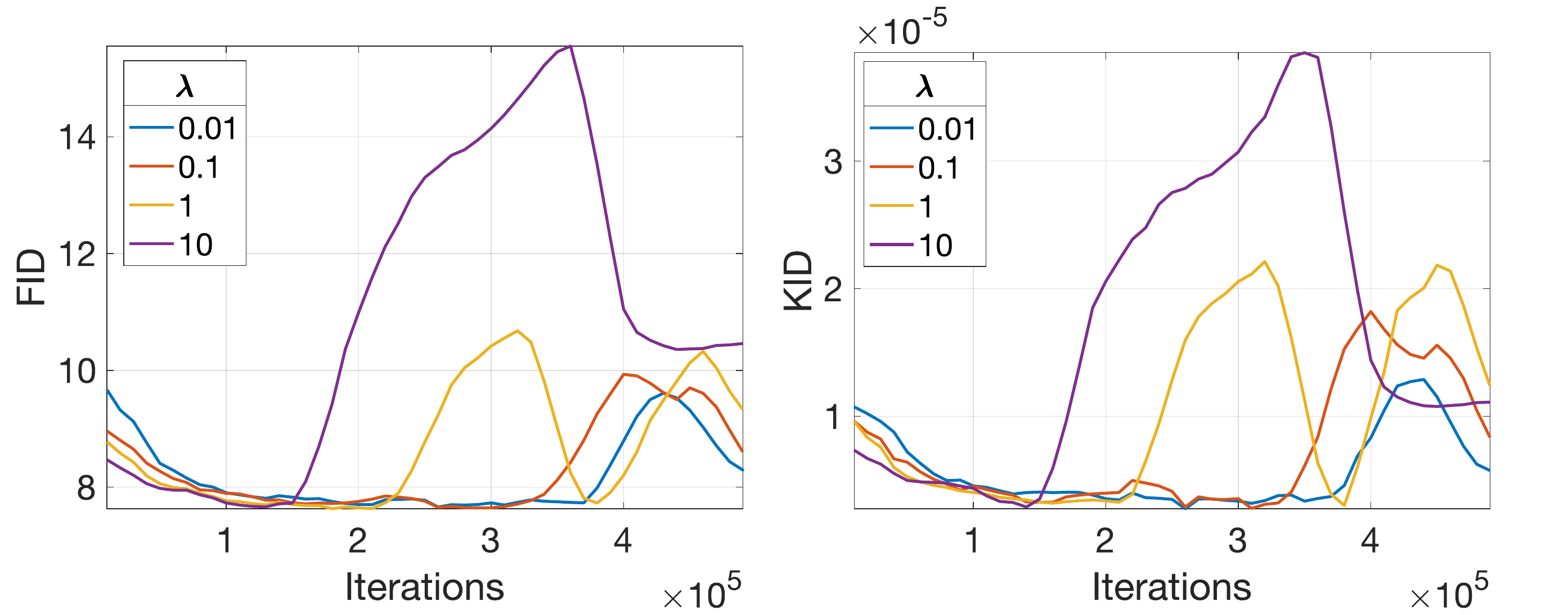}
% \vskip-0.2cm
\caption{Wasserstein GANs FID/KID scores during training for CelebA dataset.}
% \vskip-0.6cm
\label{fig2}
\end{figure}

\begin{figure*}[t]
\centering\includegraphics[width=\hsize]{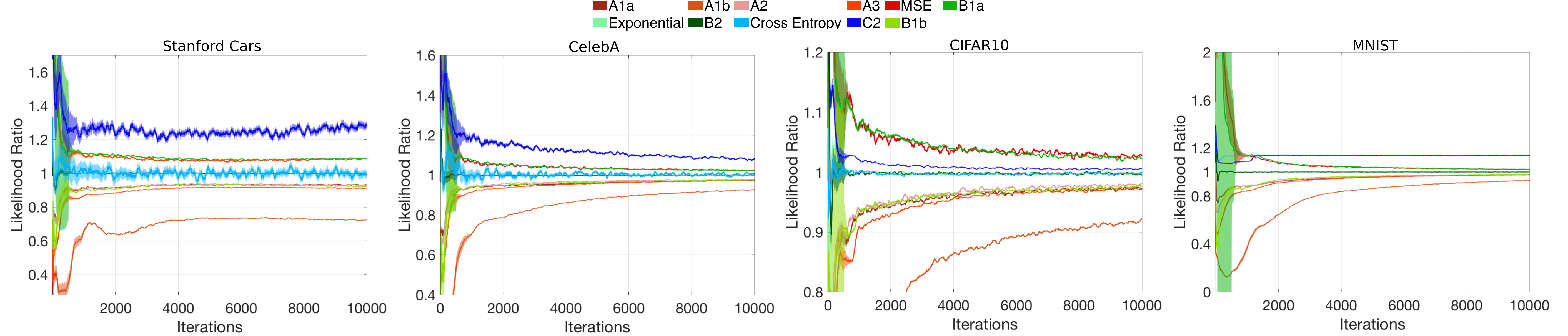}
% \vskip-0.2cm
\caption{Estimated Likelihood Ratio during training.}
\vskip-0.15cm
\label{fig1}
\end{figure*}
\begin{figure}[t]
\vskip0.2cm
\centering\includegraphics[width=\hsize]{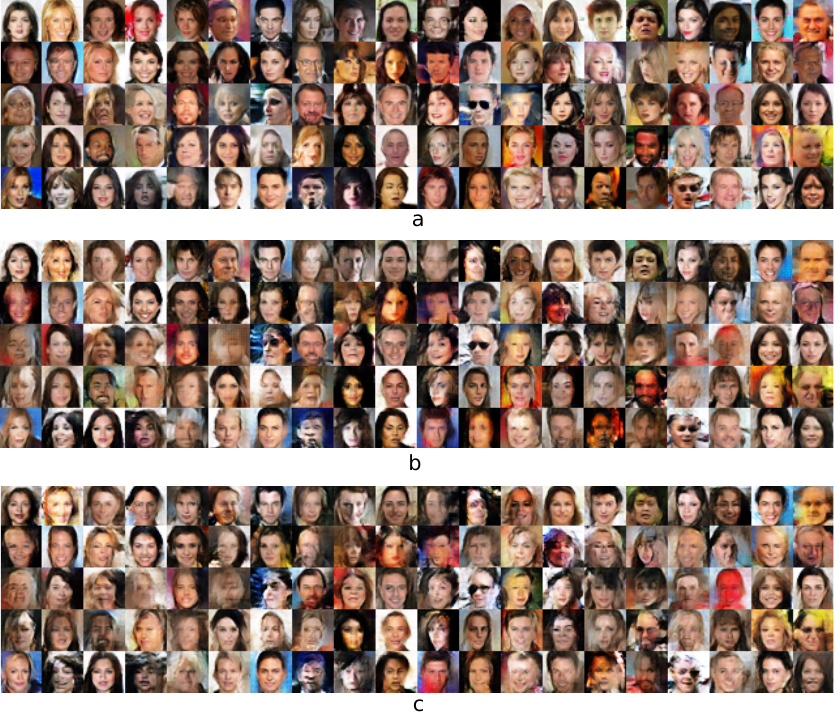}
% \vskip-0.2cm
\caption{Wasserstein GANs generated images for a - 110000, b - 210000, c - 310000 iterations for CelebA dataset.}
% \vskip-0.6cm
\label{fig3}
\vskip-0.4cm
\end{figure}
To further investigate the poor performance of the GANs mentioned above, we tried different values for the hyperparameter $\lambda$ of the maximum gradient penalty.  As in \cite{zhou2019lipschitz} we tested the values $\{0.01, 0.1, 1, 10\}$. \textit{In Table\,\ref{table:2}, we used $\lambda=10$ for all GANs since we obtained faster convergence for this value}. Indeed for $\lambda < 10$, the scores are improved for the B2, Cross-Entropy, and C2 GANs in Stanford Cars and Exponential, B2, Hinge, and Wasserstein GANs in CelebA dataset. But after continuing our training for more than $2\times10^5$ iterations, we experienced the same behavior. So the choice of $\lambda$ was not diminishing this divergent behavior but rather postponing it. Fig.\,\ref{fig2} shows an example of the previously described behavior for Wasserstein GANs with the CelebA dataset. Here for $\lambda=10$ the divergent behavior starts a little bit earlier than 150000 iterations, for $\lambda=1$ around 210000 iterations, for $\lambda=0.1$ at 300000 iterations and for $\lambda=0.01$ around 380000 iterations. Another interesting fact is that for the different values of $\lambda$, the best FID/KID scores are almost the same. So if we had to perform some early stopping in our training, it would be better to choose $\lambda=10$ and avoid the additional iterations needed for $\lambda<10$ to reach the same score. To qualitatively understand the effect of this behavior in Fig.\,\ref{fig3} we present some synthetic examples of the Generator before the algorithm starts to diverge at 110000 iterations and later at 210000 and 310000 iterations (for $\lambda=10$) where we notice the generation of blurry, distorted images.

As we mention in \ref{sec:subclasses} we can compare the different GAN losses under the same quantity, the likelihood ratio function. In other words, we can examine if they converge to the optimal value, being equal to one. \textit{Unfortunately, as we show in \ref{sec:subclasses} for Hinge and Wasserstein GANs, we cannot estimate the likelihood ratio in terms of the Discriminator function.} For the other losses, it is possible to evaluate the likelihood function from the Discriminator's output employing dataset and Generator samples during training. In Fig.\,\ref{fig1} we see the evolution of the likelihood function during the first 10000 Generator iterations where we used the training batches for its estimation. In all cases, we notice that B2, Exponential, Cross-Entropy (except MNIST) GANs converge faster to the optimum value. Interestingly additional conclusions are derived from these figures; although Cross-Entropy reaches the optimal value, it has higher variance around it than B1 and Exponential GAN (Stanford Cars and CelebA datasets). For instance, we could reduce the learning rate of the Cross-Entropy GAN to decrease the variance further.

Our simulations indicate that the Subclass A objectives A1a, A1b, A2, MSE have the best performance in terms of the computed metrics (hence image generation quality) and stability during training. Furthermore, the convergence to the optimal likelihood ratio between the dataset and the generator output can be estimated online and used as a metric to decide which GAN loss and learning rate to employ.

\section{Conclusion}
In this paper, we provided and demonstrated a straightforward methodology to determine loss functions that solve the generative adversarial problem.
Our results suggest that there is no single loss function that achieves the best performance in terms of the examined metrics for all different datasets. This performance variation among loss functions becomes evident as the increasing complexity of the datasets that convolutes the generation task is better addressed by some loss functions that clearly outperform others. Specifically, in simpler datasets, such as MNIST, the evaluated loss functions yield very similar performance, whereas, in more intricate datasets like CelebA, CIFAR-10, and Stanford Cars, performance ``gaps" between the different loss functions, and different subclasses, emerges. Our findings also propose that in every generation task, unexplored loss functions outperformed the previously proposed ones. Consequently, this function class is worth-exploring to identify new loss functions that can be used and evaluated in different applications. Our method provides a versatile tool that can be exploited in that direction.
\balance
%\bibliographystyle{IEEEtran}
%\bibliography{IEEEabrv,IEEEfull}


\end{document}